\documentclass[a4paper,twoside]{article}

\usepackage{epsfig}
\usepackage{subcaption}
\usepackage{calc}
\usepackage{amssymb}
\usepackage{amstext}
\usepackage{amsmath}
\usepackage{amsthm}
\usepackage{multicol}
\usepackage{multirow}
\usepackage{float}
\usepackage[hidelinks]{hyperref}
\usepackage{booktabs}
\usepackage{graphicx}
\usepackage{epstopdf}
\usepackage{textcomp}
\usepackage{pslatex}
\usepackage{apalike}
\usepackage[bottom]{footmisc}
\usepackage[inline]{enumitem}
\usepackage{xcolor}
\usepackage{SCITEPRESS}     

\begin{document}

\title{Combining Two Adversarial Attacks Against Person Re-Identification Systems}

\author{\authorname{Eduardo de O. Andrade\sup{1}\orcidAuthor{0000-0002-5978-9718}, Igor Garcia Ballhausen Sampaio\sup{1}\orcidAuthor{0000-0002-1890-1451}, Joris Guérin\sup{2}\orcidAuthor{0000-0002-8048-8960} and José Viterbo\sup{1}\orcidAuthor{0000-0002-0339-6624}}
\affiliation{\sup{1}Computing Institute, Federal Fluminense University, Niterói, Brazil}
\affiliation{\sup{2}LAAS-CNRS, Toulouse University, Midi-Pyrénées, France}
\email{\{eandrade, viterbo\}@ic.uff.br, igorgarcia@id.uff.br, jorisguerin.research@gmail.com}
}


\abstract{The field of Person Re-Identification (Re-ID) has received much attention recently, driven by the progress of deep neural networks, especially for image classification. The problem of Re-ID consists in identifying individuals through images captured by surveillance cameras in different scenarios. Governments and companies are investing a lot of time and money in Re-ID systems for use in public safety and identifying missing persons. However, several challenges remain for successfully implementing Re-ID, such as occlusions and light reflections in people's images. In this work, we focus on adversarial attacks on Re-ID systems, which can be a critical threat to the performance of these systems. In particular, we explore the combination of adversarial attacks against Re-ID models, trying to strengthen the decrease in the classification results. We conduct our experiments on three datasets: DukeMTMC-ReID, Market-1501, and CUHK03. We combine the use of two types of adversarial attacks, P-FGSM and Deep Mis-Ranking, applied to two popular Re-ID models: IDE (ResNet-50) and AlignedReID. The best result demonstrates a decrease of 3.36\% in the Rank-10 metric for AlignedReID applied to CUHK03. We also try to use Dropout during the inference as a defense method.}

\onecolumn \maketitle \normalsize \setcounter{footnote}{0} \vfill

\section{\uppercase{Introduction}}
\label{sec:introduction}

The amount of surveillance cameras is rising fast and could reach a market of 19.5 billion euros in the year 2023~\cite{khan2020data}. This market is related to the concept of smart cities, which aim to address sustainability themes, seeking to improve the management of risks in urban environments. 
As a result, the number of systems developed to re-identify people has increased rapidly in recent years, driven by the progress of deep neural networks~\cite{luo2019bag,kurnianggoro2017identification}. These systems are in high demand by companies and governments to address problems such as public safety, tracking people in universities and streets, behavior analysis, and even surveillance~\cite{islam2020person}. For example, this approach could help countermeasure against a terrorist offensive~\cite{shah2016multi}, such as the 9/11 attack\footnote{\href{https://www.mprnews.org/story/2021/09/10/npr-911-travel-timeline-tsa/}{https://www.mprnews.org/story/2021/09/10/npr-911-travel-timeline-tsa/}}. However, all this technological insertion ends up creating a scenario prone to software errors, hacks, malware, and other criminal activities~\cite{kitchin2019security}.

Even with the many hours of video generated by an immense number of cameras, we still need many human operators responsible for verifying incidents through observation on many screens. Automatic analysis of this data can considerably help human operators and improve the efficiency of these systems~\cite{sumari2020towards}. The research field studying this problem is called Person Re-Identification (Re-ID). It aims to distinguish specific individuals through images captured by surveillance cameras in different scenarios in the same environment~\cite{galanakis2019novelty}, such as an airport. Thanks to the large amount of data generated for Re-ID in recent years, there has been an exponential increase in publications about Re-ID systems, mostly considering deep learning solutions. For an overview of popular approaches for Re-ID, we refer the reader to the following survey~\cite{yaghoubi2021sss}.

Despite the increased performance of Re-ID models in the last decade, they are vulnerable to attacks called adversarial examples~\cite{bouniot2020vulnerability}. This attack can confuse deep neural networks, making the classification models return erroneous predictions with high confidence~\cite{goodfellow2014explaining}. An adversarial example attack on a Re-ID model can be a severe risk, such as a strike against an object detection system\footnote{\href{https://www.biometricupdate.com/201904/novel-techniques-that-can-trick-object-detection-systems-sounds-familiar-alarm}{https://www.biometricupdate.com/201904/novel-techniques-that-can-trick-object-detection-systems-sounds-familiar-alarm}}. Finding efficient attacks and countermeasures to mitigate them are active fields of research~\cite{chen2020survey}. We present a literature review about adversarial attacks in Section~\ref{sec:related_work}.

The main objective of our work is to strengthen the degeneration of the classification accuracy of a Re-ID model by combining two different types of adversarial attacks. In addition, this paper also uses a defense method for Re-ID's hardening. 
The attacks implemented and combined are \begin{enumerate*}
    \item a modification of the Fast Gradient Signed Method~\cite{goodfellow2014explaining}, known as Private Fast Gradient Signed Method (P-FGSM)~\cite{li2019scene}, and
    \item a state-of-the-art method for Re-ID, called Deep Mis-Ranking~\cite{wang2020transferable}.
\end{enumerate*}   
For the defense, we try to apply the method from~\cite{sheikholeslami2019efficient} to Re-ID, which consists in using the Dropout layers during the inference phase. As far as we know, the defense method and one of the attacks have never been used for Re-ID before.

The experiments are run using three known datasets: Duke Multi-Tracking Multi-Camera Re-Identification (DukeMTMC-ReID)~\cite{ristani2016performance}, Market-1501~\cite{zheng2015scalable} and Chinese University of Hong Kong 03 (CUHK03)~\cite{li2014deepreid}. For this work, we have the implementation of two models of Re-ID systems: AlignedReID~\cite{zhang2017alignedreid} and another system with Identification-discriminative Embedding (IDE)~\cite{zheng2016person} based on the known deep Residual Neural Network, ResNet-50~\cite{he2016deep}.

The structure of this work is in five sections. Section~\ref{sec:related_work} starts with a discussion of the different adversarial attack approaches for Re-ID present in the literature. In Section~\ref{sec:combined_attack_methods} we present the details of the two attacks used in this work. Next, Section~\ref{sec:experiments} presents the experiments performed on the implemented models and discusses the results obtained. Finally, Section~\ref{sec:conclusion} concludes this paper, describing some limitations and possible future work for this work.

\section{\uppercase{Related Work}}
\label{sec:related_work}

In 2014, there was an extensive study about adversarial examples and their effects~\cite{goodfellow2014explaining}. The authors observed that more linear models are prone to fail under attacks. The direction of perturbations was the most crucial feature in drastically altering neural network predictions. The authors also showed that adversarial examples could generalize across different models. Perturbations that are more aligned with the weight vectors of the models, learning similar functions, and training for the same tasks, facilitate generalization. Furthermore, the neural network models that are easy to optimize were easy to confuse. In 2018, another paper reviewed attack and defense approaches for deep learning models~\cite{yuan2019adversarial}, applied to tasks such as image classification, image segmentation, and object detection.

The Fast Gradient Signed Method (FGSM) approach emerged in 2014 and demonstrated how effective a simple, low-computation attack could be. It consists in adding imperceptible perturbations whose direction is the same as the gradient of the cost function concerning the data. In 2019, a variation of FGSM called Private FGSM (P-FGSM) achieved an excellent trade-off between the drop in classification accuracy and distortion of private classes~\cite{li2019scene}. The real purpose of class privacy is to protect sensitive information from images when there is an inference from a classifier. This information may include the presence of people, faces, and other content that we cannot violate. Using a ResNet-50 model and the Places365-Standard~\cite{zhou2017places} dataset, the P-FGSM authors were able to fool the classifier 94.40\% of the times in the top-5 classes with only a slight average reduction, considering three image quality measures. As far as we know, no other work in the literature used the P-FGSM in Re-ID.

The Opposite-Direction Feature Attack (ODFA) paper~\cite{zheng2018open}, implemented in 2018, used a Dense Convolution Network (DenseNet) with a depth of 121 as the victim model and another ResNet-50 model for the generation of adversarial queries. Three datasets were part of the experiments: Market-1501, Caltech University Birds-200-2011 (CUB-200-2011)~\cite{wah2011caltech} and CIFAR-10~\cite{krizhevsky2009learning}. The Market-1501 and CUB-200-2011 had better results than CIFAR-10 as ODFA handled the recovery task better. For Market-1501, the mean Average Precision (mAP) metric without the attack in a specific victim model reached an accuracy of 77.14\%~\cite{sun2018beyond}, while the attack decreased the accuracy to 21.52\% using the same model.

Another attack from 2019 has two different proposals for dealing with adversarial patterns (AdvPattern): EvdAttack and ImpAttack~\cite{wang2019advpattern}. The authors used the Market-1501 and another proprietary dataset to craft transformable patterns into adversarial clothing. The name of this proprietary dataset is Person Re-Identification in Campus Streets (PRCS). Two models were part of the experiments: a Siamese Network (A)~\cite{zheng2017discriminatively} and a ResNet-50 capable of learning the discriminative embeddings of identities (B)~\cite{zheng2016person}. For Market-1501, The mAP metric values before the application of AdvPattern are 62.7\% (model A) and 57.3\% (model B). Considering the dataset generated with EvdAttack, the authors achieved 4.4\% in model A and 4.5\% in model B. Using ImpAttack, the accuracy decreased to 9.20\% in model A and 10.9\% in model B. The adoption of PRCS with the AdvPattern approach differs from the attacks addressed in our work.

In 2020, there was an opposite approach to ODFA with the implementation of Self Metric Attack (SMA) and Furthest-Negative Attack (FNA)~\cite{bouniot2020vulnerability}. The authors performed both attacks on Market-1501 and DukeMTMC-ReID. They adopted ResNet-50 architectures using two distinct types of loss minimization: the cross-entropy (C)~\cite{xiong2019good} and the triplet loss (T)~\cite{hermans2017defense,schroff2015facenet}. The accuracy results achieved with the mAP metric for Market-1501 without the attacks were 67.22\% for T and 77.53\% for C. Using the SMA attack, there was a decrease in accuracy to 0.05\% for T and 0.26\% for C. The FNA obtained 0.05\% for T and 0.07\% for C. For the DukeMTMC-ReID dataset, the mAP results achieved without the attacks were 60.33\% for T and 67.64\% for C. Again, with the SMA attack, there was a decrease to 0.05\% for T and 0.32\% for C. The FNA obtained 0.04\% for T and 0.06\% for C.

The most important paper regarding adversarial attack approaches for this work appeared in mid-2020~\cite{wang2020transferable}. The Deep Mis-Ranking attack is responsible for most state-of-the-art results compared to our work. It is presented in details in Section~\ref{subsec:deep_mis_ranking}. However, some results obtained in our work are close to but not the same as those described in the paper. Some of the problems in implementing Deep Mis-Ranking included code errors to be corrected. The experiments were not perfectly reproducible, and results differ slightly from those initially presented in the paper, even after corrections and using models with pre-trained weights.

\section{\uppercase{Combined Attack Methods}}
\label{sec:combined_attack_methods}

There is little attention to the security risks and the impact of the attacks on Re-ID systems. This section explains the approaches used in this work: Deep Mis-Ranking, P-FGSM, and their combination.

\subsection{Deep Mis-Ranking}
\label{subsec:deep_mis_ranking}

The Deep Mis-Ranking is a formulation to disrupt the ranking prediction of Re-ID models. The main characteristic of Deep Mis-Ranking is that it has high transferability, i.e., if we implement it for dataset A, it can generalize to another dataset B. Other characteristics of Deep Mis-Ranking, include the controllability and imperceptibility of the attack~\cite{wang2020transferable}.

\begin{figure}[t]
\centering
\includegraphics[width=0.4875\textwidth]{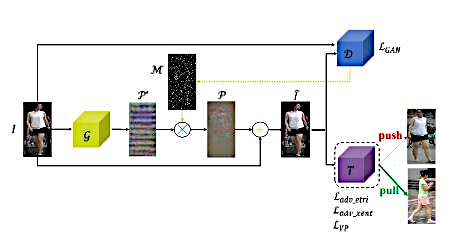}
\caption{The framework structure of the Deep Mis-Ranking attack. The main objective of the attack is to maximize the distance between the samples from the same category (pull) and minimize the distance between the samples from different categories (push). Source:~\cite{wang2020transferable}.}
\label{fig:1}
\end{figure}

Figure~\ref{fig:1} shows the visual representation of the framework. The generator $\mathcal{G}$ produces the preliminary perturbations $\mathcal{P}^{'}$ that, multiplied with the mask $\mathcal{M}$, originate the disturbances $P$ for each input image $I$. The generator $\mathcal{G}$ is a ResNet-50 architecture, and it is trained jointly with the discriminator $\mathcal{D}$ to form the general Generative Adversarial Network (GAN) structure of the framework. We commonly use this unsupervised neural network for image generation~\cite{konidaris2019generative}. $\mathcal{L}_{GAN}$ represents the GAN loss, whereas $\mathcal{L}_{adv\_etri}$, $\mathcal{L}_{adv\_xent}$, and $\mathcal{L}_{VP}$ correspond to mis-ranking, misclassification, and perception losses, respectively. The $\mathcal{T}$ represents the attacked Re-ID system and receives the adversarial image $\hat{I}$ as input.

\begin{figure}[t]
\centering
\includegraphics[width=0.4875\textwidth]{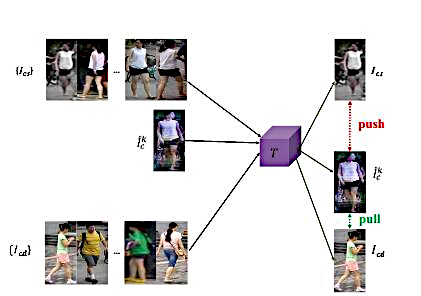}
\caption{The scheme of how the Deep Mis-Ranking attack occurs in a Re-ID system $\mathcal{T}$ concerning pairs of samples and their distances. Source:~\cite{wang2020transferable}.}
\label{fig:2}
\end{figure}

Looking more closely at $\mathcal{T}$, the inputs and outputs follow the scheme illustrated in Figure \ref{fig:2}. We aim to minimize the distance of each pair of samples from different categories $(e.g., (\hat{I}^{k}_{c}, I), \forall{I} \in \{I_{cd}\} )$ while maximizing the distance of each pair of samples from the same category $(e.g., (\hat{I}^{k}_{c}, I), \forall{I} \in \{I_{cs}\})$ to achieve a successful attack.

The Equation \ref{eq:1} corresponds to $\mathcal{L}_{GAN}$. While the $\mathcal{D}$ discriminator tries to differentiate the real images from the adversarial ones, the $\mathcal{G}$ generator tries to produce the perturbations in the input images. The expected value $\mathbb{E}_{(I_{cd}, I_{cs})}$ represents the expected conditional of $\textrm{log}\mathcal{D}_{1, 2, 3}(I_ {cd}, I_{cs})$ given $I_{cd}$ and $I_{cs}$ in the form $\mathbb{E}_{X, Y}[Y] = \mathbb{E}_{X }[\mathbb{E}_{Y}[Y | X]]$.

\begin{equation}
\begin{split}
\label{eq:1}
\mathcal{L}_{GAN} = \mathbb{E}_{(I_{cd}, I_{cs})}[\textrm{log}\mathcal{D}_{1, 2, 3}(I_{cd}, I_{cs})] + \\ 
\mathbb{E}_{\mathcal{I}}[\textrm{log}(1-\mathcal{D}_{1, 2, 3}(I, \hat{I}))]
\end{split}
\end{equation}

The first loss related to a Re-ID system $\mathcal{T}$ is $\mathcal{L}_{adv\_etri}$, represented by Equation \ref{eq:2}, where the expression $[x]_{+}$ is equal to $\textrm{max}(0, x)$. This mis-ranking loss function follows the form of a triplet loss~\cite{ding2015deep}, aiming to 
minimize the distance of mismatched pair, while maximizing the distance of the matched pair. The letter \textbf{K} represents the set of people's identities. Meanwhile, $\textrm{\textbf{C}}_{k}$ is the set of sample numbers taken from the \textit{k}-th identity of a person and $\mathcal{I}^{k}_{c}$ are the \textit{c}-th images of the $k$ identity in a \textit{mini-batch}. The L2 norm used as a distance metric is represented by $|| \cdot ||_{2}$ and $\Delta$ is a margin threshold.

\begin{equation}
\begin{split}
\label{eq:2}
\mathcal{L}_{adv\_etri} = \sum_{k = 1}^{\textrm{\textbf{K}}}\sum_{c = 1}^{\textrm{\textbf{C}}_{k}}[\underset{\underset{\underset{c_{d} = 1 ... \textrm{\textbf{C}}_{j}}{j = 1 ... \textrm{\textbf{K}}}}{j \neq k}}{\textrm{max}} || \mathcal{T}(\hat{I}^{k}_{c}) - \mathcal{T}(\hat{I}^{j}_{c_{d}})||^{2}_{2} - \\ \underset{c_{s} = 1 ... \textrm{\textbf{C}}_{k}}{\textrm{min}} || \mathcal{T}(\hat{I}^{k}_{c}) - \mathcal{T}(\hat{I}^{j}_{c_{s}})||^{2}_{2} + \Delta]_{+}
\end{split}
\end{equation}

Another loss present in the framework is $\mathcal{L}_{adv\_xent}$ for non-targeted attack (Equation~\ref{eq:3}), where $\mathcal{S}$ denotes the softmax function and $\delta$ is the Dirac delta. The term $\upsilon$ is the smoothing regularization, where $\upsilon = [\frac{1}{K-1}, ..., 0, ..., \frac{1}{K-1}] $, where $\upsilon$ is always equal to $\frac{1}{K-1}$ except in the case where $k$ is the ground-truth ID (\textrm{\textbf{K}} is the set including each \textit{k-th} person ID). The arg min preceded by $\mathbb{I}$ represents the case in which we have the return of the indices of the minimum values of an output probability vector, indicating the least likely class (similar to the numpy.argmin function present in the NumPy library of Python).

\begin{equation}
\label{eq:3}
\mathcal{L}_{adv\_xent} = -\sum_{k = 1}^{\textrm{\textbf{K}}}\mathcal{S}(\mathcal{T}(\hat{\mathcal{I}})_{k}((1-\delta)\mathbb{I}_{\textrm{arg min} \mathcal{T}(\mathcal{I})_{k}} + \delta_{\upsilon_{k}})
\end{equation}

In order to improve the visual quality for $\mathcal{T}$ and prevent the attack from being detected by humans, we have the Equation \ref{eq:4} corresponding to the perception loss $\mathcal{L}_{VP }$. The formulation of this loss function originates from an approach to the structural similarity image quality paradigm~\cite{wang2003multiscale}. The comparison measures of contrast ($c_{j}$) and structure ($s_{j}$) on the \textit{j}th scale are calculated by $c_{j}(\mathcal{I}, \mathcal {\hat{I}}) = \frac{2\sigma_{\mathcal{I}}\sigma_{\mathcal{\hat{I}}} + C_{2}}{\sigma^{2}_{ \mathcal{I}}+\sigma^{2}_{\mathcal{\hat{I}}} + C_{2}}$ and $s_{j}(\mathcal{I}, \mathcal{\hat {I}}) = \frac{\sigma_{\mathcal{I}\mathcal{\hat{I}}} + C_{3}}{\sigma_{\mathcal{I}}\sigma_{\mathcal{\ hat{I}}} + C_{3}}$, where $\sigma_{x}$ is the standard deviation, 
$\sigma^{2}_{x}$ is the variance and $\sigma_{xy}$ of covariance. The level of the scales is represented by $L$, where $\alpha_{L}$, $\beta_{j}$ and $\gamma_{j}$ are the factors that help to re-weight the contribution of each component. Finally, we have the luminosity measure ($l$) calculated by $l_{L}(\mathcal{I}, \mathcal{\hat{I}}) = \frac{2\mu_{\mathcal{I} }\mu_{\mathcal{\hat{I}}} + C_{1}}{\mu^{2}_{\mathcal{I}}+\mu^{2}_{\mathcal{\hat{ I}}} + C_{1}}$, where $\mu_{x}$ is the mean form.

\begin{equation}
\label{eq:4}
\mathcal{L}_{VP} = [l_{L}(\mathcal{I}, \mathcal{\hat{I}})]^{\alpha_{L}} \cdot \prod^{L}_{j = 1}[c_{j}(\mathcal{I}, \mathcal{\hat{I}})]^{\beta_{j}}[s_{j}(\mathcal{I}, \mathcal{\hat{I}})]^{\gamma_{j}}
\end{equation}

The $\mathcal{M}$ mask determines the number of target pixels to attack. After multiplying the preliminary perturbation $\mathcal{P}^{'}$ with the mask $\mathcal{M}$, we have the final perturbation $\mathcal{P}$ with a controlled number of pixels enabled to maintain discretion from the attack. The function Gumbel softmax~\cite{jang2016categorical} is responsible for choosing pixels from all possibilities. The generalization capacity of Deep Mis-Ranking is its main advantage. It is possible to use it with different Re-ID systems and efficiently in black-box scenarios.

\begin{figure}[t]
\centering
\includegraphics[width=0.4825\textwidth]{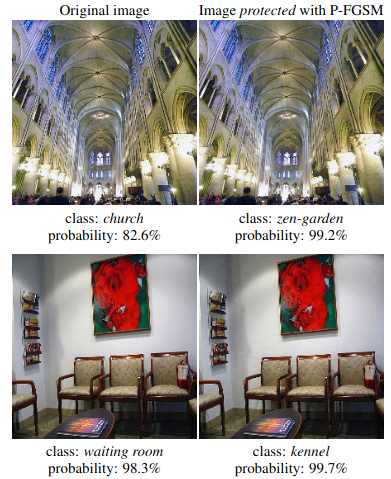}
\caption{The two images on the left side represent the true class. On the right side, we have the distortions that make them human-imperceptible and give a high misclassification rate to the models. Source:~\cite{li2019scene}.}
\label{fig:3}
\end{figure}

\subsection{Private Fast Gradient Signed Method} \label{subsec:private_fast_gradient_signed_method}

The design of the P-FGSM aims to ``protect'' the data of an image through directed distortions that make it difficult to infer a classifier. The purpose of this approach is to maintain usefulness for social media users. P-FGSM is based on the FGSM attack already used in Re-ID and includes a limitation on the probability that automatic inference can expose the true class of a distorted image. This limitation may include even more disturbances that mislead models~\cite{li2019scene}.

Figure \ref{fig:3} shows an example of two images after executing P-FGSM. The most significant difference between FGSM and other variants of this attack is irreversibility, i.e., the random selection of the target class among the subset of classes that do not contain the protected class. The target class and other classes can denote people's different identities in a Re-ID dataset.

The class adapted as an adversarial example $\widetilde{y}$ works as a function of the classification probability vector $\textbf{p}$. So, $\textbf{p}$ being equal to the vector of features for classification, we have $\textbf{p'}$ which contains the elements of $\textbf{p}$ in descending order, $\textbf{p'} = (p'_{1}, ..., p'_{D})$, where $D$ represents the scene classes. Equation \ref{eq:5} corresponds to the random choice of $\widetilde{y}$ from the subset of classes whose cumulative probability exceeds a threshold $\sigma$ in the interval [0,1], in which R is the function that randomly picks one class label $y_{j}$ from the input set. 

\begin{equation}
\label{eq:5}
\widetilde{y} = R\left(\left\{y_{j}:\sum^{j-1}_{i = 1}\textbf{p'}_{i} > \sigma\right\}\right)
\end{equation}

Lastly, in Equation \ref{eq:6}, we have the generation of the protected image $\dot{x} = \dot{x}_{N}$ for $N$ iterations, starting from $\dot{x} _{0} = x$ to a maximum number of iterations aiming to increase the probability of predicting $\widetilde{y}$. We can represent the cost function by $J_{M}$, and it is used in training to estimate the $\theta$ parameters of the classifier $M$. The $\epsilon$ represents the measure of the maximum magnitude of the adversarial
perturbation and $\nabla$ is, in this case, the gradient vector that is related to the image $x$.

\begin{equation}
\label{eq:6}
\dot{x}_{N} = \dot{x}_{N-1} - \epsilon sign(\nabla_{x}J_{M}(\theta, \dot{x}_{N-1}, \widetilde{y}))
\end{equation}

\section{\uppercase{Experiments}}
\label{sec:experiments}

We used a computer with a 2.9 GHz Intel Xeon processor and 16 GB 2400 MHz DDR4 of RAM for evaluation purposes using the GPU Nvidia Quadro P5000. The datasets used where DukeMTMC-ReID~\cite{ristani2016performance}, Market-1501~\cite{zheng2015scalable}, and CUHK03~\cite{li2014deepreid}. DukeMTMC-ReID had 16,522 images (bounding boxes) with 702 identities for training and 19,889 images with 702 other identities for testing. We used 2228 bounding boxes to correctly identify the test identities considering the query set. For Market-1501, the composition was 12,936 images of 751 identities for training and 19,281 images of 750 identities for testing. We selected 3368 bounding boxes for the query set. Finally, CUHK03 comprised 7365 images of 767 identities for training and 6732 images of 700 identities for testing, and the query set contained 1400 images. It is important to mention that we neglected some ``junk images'' from Market-1501 in our testing set. These images were neither good nor bad considering the Deformable Part Model (DPM) bounding boxes; they could hinder more than help, making no difference in the re-identification process and accuracy. The DPM is a pedestrian detector employed instead of the hand-cropped boxes. We also did not use some images from CUHK03 that we could not read from the MATLAB file that composes the dataset.

\begin{table*}[t]
\centering
\caption{The results (in percent) with and without combined attacks for the chosen models and datasets}
\label{tab:1}
\begin{tabular}{@{}cc|cccc|cccc@{}}
\toprule
\multirow{2}{*}{Dataset} & \multirow{2}{*}{Method} & \multicolumn{4}{c|}{IDE} & \multicolumn{4}{c}{AlignedReID} \\
 & & mAP & R-1 & R-5 & R-10 & mAP & R-1 & R-5 & R-10 \\ \midrule
DukeMTMC-ReID & \begin{tabular}{@{}c@{}}No Attacks \\ Deep Mis-Ranking \\ P-FGSM \\ Combined Attacks\end{tabular}   &   \begin{tabular}{@{}c@{}}58.14 \\ 4.68 \\ 56.06 \\ 4.71\end{tabular}   &   \begin{tabular}{@{}c@{}}76.53 \\ 5.16 \\ 75.45 \\ 5.25\end{tabular}   &   \begin{tabular}{@{}c@{}}86.76 \\ 8.71 \\ 86.54 \\ 9.87\end{tabular}   &   \begin{tabular}{@{}c@{}}89.99 \\ 11.00 \\ 90.08 \\ 11.98\end{tabular}  & \begin{tabular}{@{}c@{}}69.75 \\ 3.12 \\ 67.05 \\ \textbf{3.09}\end{tabular} & \begin{tabular}{@{}c@{}}82.14 \\ 3.23 \\ 81.82 \\ 3.77\end{tabular}   &   \begin{tabular}{@{}c@{}}91.65 \\ 6.01 \\ 91.16 \\ 7.00\end{tabular}   &   \begin{tabular}{@{}c@{}}94.43 \\ 7.99 \\ 93.85 \\ 8.75\end{tabular} \\  \hline 
Market-1501 & \begin{tabular}{@{}c@{}}No Attacks \\ Deep Mis-Ranking \\ P-FGSM \\ Combined Attacks\end{tabular}   &   \begin{tabular}{@{}c@{}}61.13 \\ 4.30 \\ 58.08 \\ \textbf{4.24}\end{tabular}   &   \begin{tabular}{@{}c@{}}80.85 \\ 3.98 \\ 79.33 \\ \textbf{3.95}\end{tabular}   &   \begin{tabular}{@{}c@{}}91.89 \\ 8.88 \\ 91.27 \\ 9.38\end{tabular}   &   \begin{tabular}{@{}c@{}}94.83 \\ 12.23 \\ 94.12 \\ 12.83\end{tabular} & \begin{tabular}{@{}c@{}}79.10 \\ 2.58 \\ 76.84 \\ \textbf{2.44}\end{tabular}   &   \begin{tabular}{@{}c@{}}91.83 \\ 1.84 \\ 91.12 \\ 1.96\end{tabular}   &   \begin{tabular}{@{}c@{}}96.97 \\ 4.22 \\ 96.82 \\ 4.54\end{tabular}   &   \begin{tabular}{@{}c@{}}98.13 \\ 6.29 \\ 98.25 \\ 6.71\end{tabular}  \\    \hline    
CUHK03 & \begin{tabular}{@{}c@{}}No Attacks \\ Deep Mis-Ranking \\ P-FGSM \\ Combined Attacks\end{tabular}   &   \begin{tabular}{@{}c@{}}24.54 \\ 0.77 \\ 19.53 \\ \textbf{0.57}\end{tabular}   &   \begin{tabular}{@{}c@{}}24.93 \\ 0.29 \\ 21.14 \\ \textbf{0.07}\end{tabular}   &   \begin{tabular}{@{}c@{}}43.29 \\ 1.00 \\ 35.79 \\ \textbf{0.79}\end{tabular}   &   \begin{tabular}{@{}c@{}}51.79 \\ 1.71 \\ 45.64 \\ 1.71\end{tabular} & \begin{tabular}{@{}c@{}}59.65 \\ 2.19 \\ 50.05 \\ \textbf{1.76}\end{tabular}   &   \begin{tabular}{@{}c@{}}61.50 \\ 1.36 \\ 53.50 \\ \textbf{1.14}\end{tabular}   &   \begin{tabular}{@{}c@{}}79.43 \\ 2.50 \\ 75.07 \\ \textbf{1.93}\end{tabular}   &   \begin{tabular}{@{}c@{}}85.79 \\ 4.36 \\ 82.21 \\ \textbf{3.36}\end{tabular} \\    \bottomrule
\end{tabular}
\end{table*}

The implemented models were IDE (ResNet-50)~\cite{he2016deep} and AlignedReID~\cite{zhang2017alignedreid}. In addition to the Deep Mis-Ranking~\cite{wang2020transferable} and P-FGSM~\cite{li2019scene} that we use as a combined attack against the models that characterize the Re-ID systems, we also implemented the Dropout at inference as a defense method. As far as we know, this defense method was not implemented yet for Re-ID systems.

Table~\ref{tab:1} shows the results using the metrics mean Average Precision (mAP), Rank-1 (R-1), Rank-5 (R-5), and Rank-10 (R-10) for the experiments with and without the combined attacks. Considering the combined attacks, we implemented one attack after the other, using P-FGSM first. There was no significant difference in changing the order of the attacks, and we used the same pre-trained weights from the Deep Mis-Ranking work\footnote{\href{https://github.com/whj363636/Adversarial-attack-on-Person-ReID-With-Deep-Mis-Ranking}{https://github.com/whj363636/Adversarial-attack-on-Person-ReID-With-Deep-Mis-Ranking}}.

Looking again at Table~\ref{tab:1}, if we compare the results without attacks and with Deep Mis-Ranking only, there are differences concerning the original paper. For IDE (ResNet-50), for instance, the results without attacks are equal in our experiments using the exact implementation and different with Deep Mis-Ranking. We used the same split for training and test sets. So, this difference could be about the dataset and its samples because it is no longer available on the official repository site or even something related to the available pre-trained weights.

We tried to strengthen the combined attacks' decrease in the classification results. This decrease occurred more times with the CUHK03 dataset, as shown in bold at Table~\ref{tab:1}. However, if we look at all the datasets and models, there are more times with a slight increase in the considered metrics, but this rise seems less critical than decreasing, even more, the results compared to the Deep Mis-Ranking attack.

\begin{figure}[t]
\centering
\includegraphics[width=0.50\textwidth]{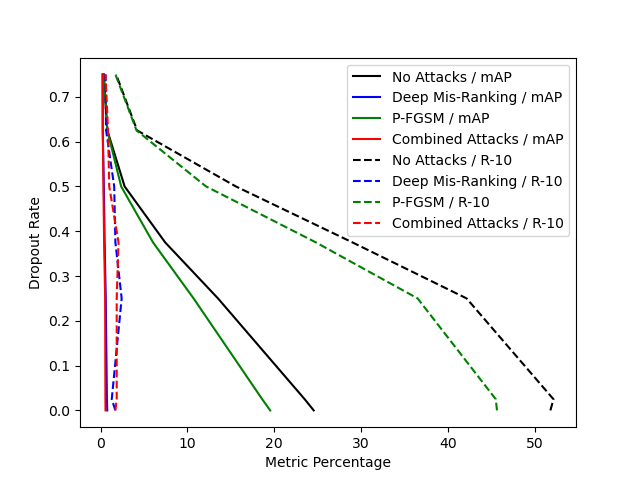}
\caption{The Dropout rate \& metric percentage with and without attacks for the mAP and Rank-10 (R-10) metrics with the IDE model and CUHK03 dataset.}
\label{fig:4}
\end{figure}

Furthermore, we used Dropout during the inference as a defense method. We expected a good trade-off for the Re-ID system against adversarial examples, changing some loss in identification results without attacks for a considerable gain in decreasing the loss in identification results, considering the attacks. Nonetheless, unlike in other cases, we did not get significant results using that method for the Re-ID systems. We can see the results of this trial in Figure~\ref{fig:4} for the mAP and Rank-10 metrics with the IDE model and CUHK03 dataset.

The Dropout behavior in Figure \ref{fig:4} illustrates the insignificant gain as a defense method. We used rate values for Dropout from 0.025 to 0.75, and the best increase was in the R-10 metric against the Deep Mis-Ranking attack, with a rate of 0.25, improving from 1.71\% to 2.73\%. Meanwhile, we have a decrease in the mAP metric using the same rate from 0.77\% to 0.60\%, which does not pay off. For the other model and datasets, the results were not good enough too.

Finally, the Dropout during the inference considered all the hidden layers of the two models. The time for running the experiments on the testing set for IDE (ResNet-50) model and DukeMTMC-ReID dataset was approximately 4 minutes. For the Market-1501 dataset, 4 minutes and 30 seconds. The CUHK03 dataset spent nearly 1 minute and 30 seconds of the execution time. Considering the AlignedReID model and DukeMTMC-ReID dataset, we finished in approximately 8 minutes. For the Market-1501 dataset, it was 11 minutes. Lastly, the CUHK03 dataset spent 2 minutes and 30 seconds.

\section{\uppercase{Conclusion}}
\label{sec:conclusion}

In this work, we proposed the combination of two adversarial attacks against Re-ID systems. As far as we know, one of the attacks, the P-FGSM, was never implemented before for this kind of system. More than that, we used Dropout during the inference as a countermeasure for the considered attacks.

We used three datasets and two models with the best results and among the most used ones for the experiments. Our tests aimed to increase the obstacles even further for Re-ID with the combination of the attack methods. These tests strengthen the decrease in the classification results in some cases. However, the proposed countermeasure did not perform well against the attacks.

There were limitations related to the accessible data and unexpected results considering the already available attack implementations. However, we pretend to continue exploring this problem concerning adversarial attacks and Re-ID systems. We also hope that combining different attack and defense methods can be an approach for our future work and other works.

\section*{\uppercase{Acknowledgement}}
\label{sec:Acknowledgements}

The \textit{Coordenação de Aperfeiçoamento de Pessoal de Nível Superior} (CAPES), \textit{Conselho Nacional de Desenvolvimento Científico e Tecnológico} (CNPq), and \textit{PrimeUp Soluções de TI LTDA} financed part of this work.

\bibliographystyle{apalike}
{\small
\bibliography{example}}

\end{document}